# Stack of discriminative autoencoders for multiclass anomaly detection in endoscopy images

Mohammad Reza Mohebbian, Khan A. Wahid, and Paul Babyn

M.R. Mohebbian, K.A. Wahid, are with Department of Electrical and Computer Engineering, University of Saskatchewan, Saskatoon, SK S7N 5A9, Canada.
P.Babyn is with Saskatchewan Health Authority, Saskatoon, SK S7K 0M7, Canada.

*Abstract*— Wireless Capsule Endoscopy (WCE) helps physicians examine the gastrointestinal (GI) tract noninvasively. There are few studies that address pathological assessment of endoscopy images in multiclass classification and most of them are based on binary anomaly detection or aim to detect a specific type of anomaly. Multiclass anomaly detection is challenging, especially when the dataset is poorly sampled or imbalanced. Many available datasets in endoscopy field, such as KID2, suffer from an imbalance issue, which makes it difficult to train a high-performance model. Additionally, increasing the number of classes makes classification more difficult. We proposed a multiclass classification algorithm that is extensible to any number of classes and can handle an imbalance issue. The proposed method uses multiple autoencoders where each one is trained on one class to extract features with the most discrimination from other classes. The loss function of autoencoders is set based on reconstruction, compactness, distance from other classes, and Kullback-Leibler (KL) divergence. The extracted features are clustered and then classified using an ensemble of support vector data descriptors. A total of 1,778 normal, 227 inflammation, 303 vascular, and 44 polyp images from the KID2 dataset are used for evaluation. The entire algorithm ran 5 times and achieved F1-score of 96.3 ± 0.2% and 85.0 ± 0.4% on the test set for binary and multiclass anomaly detection, respectively. The impact of each step of the algorithm was investigated by various ablation studies and the results were compared with published works. The suggested approach is a competitive option for detecting multiclass anomalies in the GI field.

## 1. INTRODUCTION

Endoscopy is the gold standard for examining the GI tract and is critical for detecting GI illnesses early [1]. Traditional endoscopic techniques such as colonoscopy and gastroscopy are invasive, but they enable real-time video inspection and can detect a variety of diseases such as polyposis syndromes, esophagitis, and ulcerative colitis [2]. On the other hand, Wireless Capsule Endoscopy (WCE) provides a noninvasive way for GI imaging of regions that are not accessible using traditional methods and it is less painful for the patients [3].

The main reason for analyzing recorded GI videos is detecting anomalies. Gastroenterologists are unable to locate necessary diagnostically significant frames due to the lengthy manual inspection process owing to the large volume of video data and intrinsic redundancy. For example, research reveals that the accuracy of gastroenterologists diagnosing a small polyp (less than 1 cm) is roughly 76% [4]. A clinician requires two hours on average to examine about 50,000 images and issue a diagnosis report for a specific patient, according to [5]. Therefore, using a computer-aided diagnosis system with image processing and machine-learning algorithms may save costs and time as well as reduce human errors [6].

The literature is divided into three types of anomaly detection. Firstly, various studies have targeted specific anomaly detection, such as bleeding, which is a binary classification. For instance, Li and Meng [7] showed that a combination of colour and texture traits is more successful than either colour or texture feature alone in precisely identifying polyps. Bernal *et al.* [8] utilized an inpainting diffusion method in combination with an energy map to locate polyps on a publicly available dataset [9] and acquired 84.2% accuracy. Another private dataset was used by Klare *et al.* [10] with software named APDS for polyp identification. They deployed endoscopists to evaluate quantitative results and found 85.3% accuracy. Hassan *et al.* [11] used a private dataset with Medtronic software for polyp detection and could achieve 82% accuracy. Gulati *et al.* [12] used a convolutional neural network (CNN) and achieved 90% recall and 63% specificity for polyp identification. Zhang *et al.* [13] applied transfer learning for feature extraction on a private dataset and used SVM for polyp detection, which achieved 85.9% accuracy and an 87% F1-score. The KID1 dataset [14] was utilized by Georgakopoulos *et al.* [15] for inflammatory detection. They used CNN architecture and achieved 90.2% accuracy. As mentioned earlier, most of the published research in endoscopy comes from this category. However, a more generalized model can be trained to detect multiple anomalies.

Secondly, there is some other research aimed at detecting anomalies in binary forms and as a general concept. In other words, classification is applied to classify normal and abnormal images, while the type of abnormality is not specified. Jain *et al*. [16] combined KID1 and KID2 and used a random forest-based ensemble classifier with fractal features. They achieved 85% accuracy and an 84% F1-score. Diamanti *et al*. [17] used a modified CNN method, called look-behind fully CNN, on the KID2 dataset for anomaly detection. They achieved 88.2% accuracy using 10-fold cross validation. Vasilakakis *et al*. [18] used a modified CNN for detecting anomalies on the KID2 dataset with a binary approach and achieved 90.0% AUC. This category, like the previous one, only focuses on binary classification.

Thirdly, there is less research on targeted multiclass anomaly detection [19, 20]. Mohammed *et al*. [19] used residual Long Short-Term Memory architecture for classifying 14 different anomalies and achieved a 55.0% F1-score. They showed that as the number of classes grows, the problem becomes more complicated and performance reduces. This issue is worse in clinical data, where an imbalance issue is one of the most common issues. Nawarathna *et al*. [20] used textons dictionary with KNN to classify images as erythema, blood (bleeding), ulcer, erosion, polyp, or normal, and achieved 91.1% accuracy. Despite advances in deep learning and machine learning approaches, multiclass anomaly detection is still new, and more research can be done.

One of the most important criteria for successful classification is the selection of specific features capable of capturing the internal structure of the data. Deep learning has shown promising performance in extracting features from clinical data leading to high performance classification [21]; however, it is more suitable for large data that is well sampled and has a roughly similar sampling number in each class. Autoencoders are a type of deep learning approach that can extract features in an unsupervised fashion. In other words, they do not need a balanced dataset for feature extraction, however, it is not guaranteed they can extract discriminant features to help in classification especially when there are not enough training examples in some classes. Moreover, when the number of classes increases, the sophistication of the problem increases. However, dividing the problem into multiple simple sections, which is known as decomposition strategy [22], can help to overcome this issue by solving simpler parts. The main contribution of this work is introducing an algorithm for extracting features and multiclass classification that can work well when data is poorly sampled or imbalanced.

We extracted features using multiple autoencoders where each one is trained to generate features that are specific to ones of the classes. Features are extracted based on distance metric learning and reconstruction loss in supervised fashion. Extracted features are then clustered to small parts and for each cluster, a one-class classification (OCC) algorithm is trained. The outcomes of the OCCs are combined and ensembled using XGBoost for predicting anomalies. Various ablation studies are performed to show the impact of each step of the algorithm, such as the effect of OCC compared to other classifiers, and finally the proposed method is compared with other popular techniques, such as the transfer-learning approach.

This paper is organized as follows: the next section presents information about images and the formulation of methods used in this study; section 3 provides the results of the proposed method; the discussion is provided in section 4; and the conclusion is the last section.

## 2. MATERIALS AND METHODS

### 2.1 Dataset

This study uses the KID2 dataset, which contains images captured by MiroCam capsule endoscope with 360×360 pixels resolution [23]. From KID2, 227 images of inflammatory abnormalities (aphthae, cobblestone mucosa, ulcers, luminal stenosis, mucosal/villous oedema, and mucosal breaks with surrounding erythema and/or fibrotic strictures), 44 images of polyposis abnormalities (lymphoma, lymphoid nodular hyperplasia, and Peutz-Jeghers polyps), 303 images with vascular anomalies (small bowel angiectasias and blood in the lumen), and 1,778 normal images were acquired from the esophagus, stomach, small intestine, and colon. A detailed description is provided in Table 1.

**TABLE 1**. Images used from each dataset

| Dataset | Type | | No. of images | Description |
|---|---|---|---|---|
| KID2 | Normal | | 1,778 | without anomaly and without specified locations |
| | Anomaly | Inflammation | 227 | aphthae, cobblestone mucosa, ulcers, luminal stenosis, mucosal/villous oedema, and mucosal breaks with surrounding erythema and/or fibrotic strictures |

| | | |
|---|---|---|
| Vascular | 303 | small bowel angiectasias and blood in the lumen |
| Polyps | 44 | lymphoma, lymphoid nodular hyperplasia, and Peutz-Jeghers polyps |

## 2.2 Proposed Method

The block diagram of the proposed method is shown in Figure 1. The proposed algorithm has two main stages—feature extraction and classification. The feature extraction is performed using latent vector generated by training multiple autoencoders that each one is sensitive to one class. The classification part is a combination of unsupervised clustering and OCCs which are used like feature transformer. The results of OCCs are ensembled using the XGboost classifier for detecting anomalies. Each step is discussed in the next subsections.

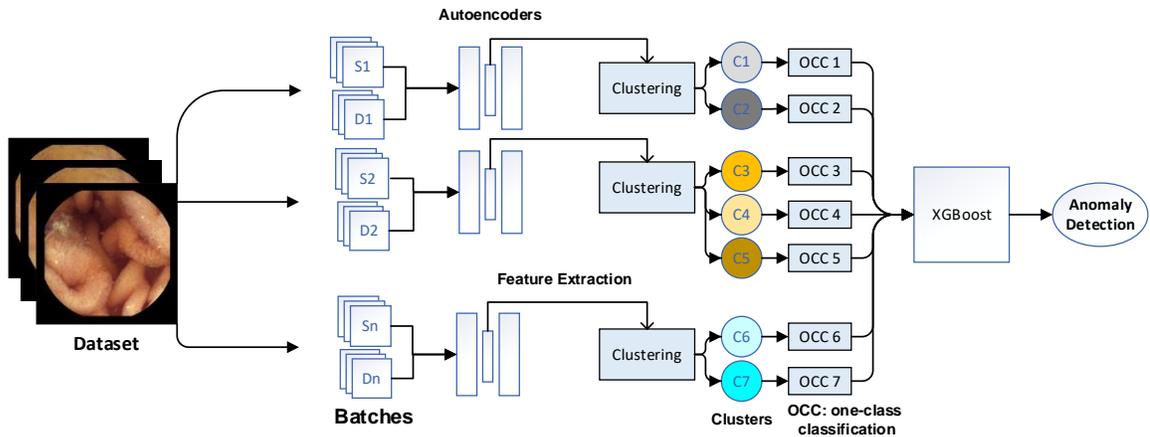

**Figure 1.** The proposed framework based on one-class classification. $Sn$: images from class n in one batch. $Dn$: images that are different than class n; $C1$: Cluster number 1; OCC: one-class classification.

### 2.2.1 Autoencoder architecture

For extracting features from images, the variational autoencoder concept is utilized. Three Convolution Layers (Conv2D) are applied on the RGB image along with the Leaky Rectified Linear Unit activation (LeakyReLU) function [24], batch normalization and MaxPooling layers. Three encoding layers have filter sizes 8, 16 and 32, and kernel size $(3 \times 3)$. Three MaxPooling layers in an encoder are used to reduce the feature size and subsequently have sizes $(3 \times 3)$, $(4 \times 4)$ and $(8 \times 8)$. Extracted filters are flatten and feed to two Dense layers with size 256, which creates the average and standard deviation for generating a Gaussian sample. The sampling procedure must be expressed in such a way that the error can be propagated backwards through the network. Hence, the reparameterization trick is used to make gradient descent practical despite the random sampling that occurred halfway through the structure. A similar architecture is used for decoding, however, instead of the MaxPooling layer it uses the Upsampling layer. Figure 2 shows the architecture of the proposed autoencoder.

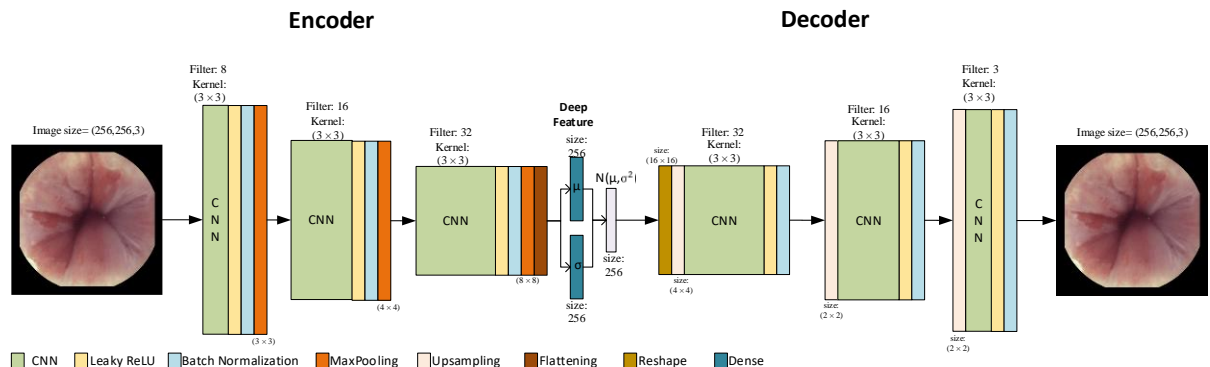

**Figure 2.** The architecture of the base autoencoder model for feature extraction. This part is shown as autoencoder in Figure 1.

### 2.2.2 Training discriminant autoencoders

A desirable feature quality is that different images from the same class have similar feature representations while images from a different class have a different representation. As a result, a set of features derived from images belonging to one class should be compactly positioned in the feature space far from other features belonging to other classes. However, endoscopy images have high similarity and abnormal regions are usually small which is hard to capture in the model. Therefore, the latent vectors (deep feature) of different classes, acquired from proposed autoencoder, were not discriminated enough. In this regard, $N$ biased models ($N$ is equal to the number of classes) are trained to increase the Euclidean distance of latent vectors from different classes (distance loss) and decrease the distance of features inside the target class (compactness loss). The Kullback-Leibler (KL) divergence and reconstruction loss based on mean square error was also used, since those are used in the architecture of many autoencoders [25]. The KL loss measure of divergence between distribution and reconstruction loss guarantees that extracted features are descriptive enough to reconstruct the image from them. The loss function is defined in equation 1:

$$l_{total} = (1 - \lambda_k - \lambda_c - \lambda_d)l_r + \lambda_c l_c + \lambda_d l_d + \lambda_k l_k \quad (1)$$

$$0 \leq \lambda_c, \lambda_d, \lambda_k \leq 1, \sum \lambda_i = 1$$

Where, $l_c$ is the compactness loss, $l_d$ is the distance loss, $l_k$ is KL divergence loss and $\lambda$ is the weighting parameter. We used $\lambda_c = 0.25$, $\lambda_d = 0.25$ and $\lambda_k = 0.25$ to assign the same weight on each objective.

If the problem is the $N$-class classification, $N$ autoencoder can be trained to optimize loss function (1). Each autoencoder gets two batches of data; one batch contains images from a specific class, another batch contains images with a combination of other classes. Compactness loss is the mean squared intra-batch distance inside a given batch containing one class and it aims to reduce covariance of data. Although any possible distance metric can be used, Euclidean distance is used in this research. Distance loss is defined as an average of Euclidean distance between latent vectors acquired from two batches. The following equations define compactness loss and distance loss:

$$l_c = \frac{1}{nk} \sum_{i=1}^{n} (\mu_i - m_i)^T (\mu_i - m_i) \quad (2)$$

$$l_d = \frac{1}{n^2} \sum_{j=1}^{n} \sum_{i=1}^{n} \|\mu_i - \mu_j\|_2 \quad (3)$$

Where, $\mu_i = (x_1, x_2, \ldots x_k)$ is the deep feature extracted by an autoencoder in a batch of target class, $\mu_i = (x_1, x_2, \ldots x_k)$ is the deep feature extracted by an autoencoder in a batch of other classes, $m_i$ is the average of the deep feature $\mu_i$ and $n$ is the batch size, which is set to 8 in this research. Figure 3 visualizes the concept of creating $N$ feature extraction model. Each model is trained to discriminate one specific class from others and is trained using Adam optimizer [26] for 100 epochs.

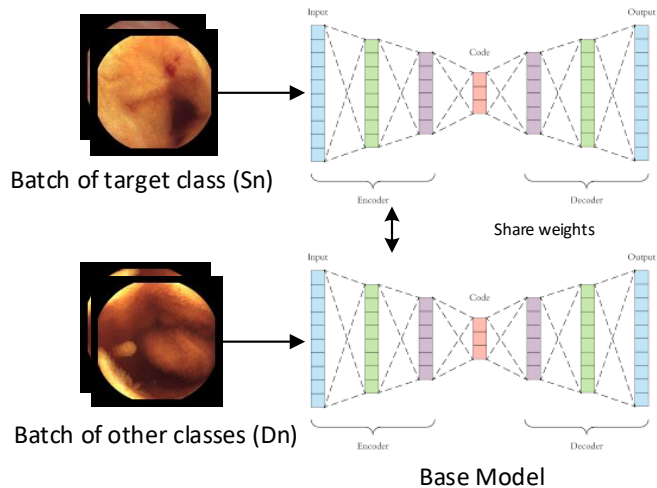

**Figure 3.** Training a autoencoder for extracting discriminant deep feature for target class.

### 2.2.3 Classification

In this paper, two different classifiers are trained and evaluated. First, all data are labeled as normal and abnormal, and one model is trained for anomaly detection in binary form. For the second model, a multiclass classification approach is utilized. In each case, extracted features from trained autoencoders are used for training classifier.

Various classification algorithms can be adopted; however, using an ensemble of OCCs, which are trained on clusters from feature space results in better performance. First, the OCC can concentrate on only one class and conform to the corresponding target class. Therefore, the unique properties of a class can be captured while preventing over-fitting at the same time. This also helps improve generalization [27]. Secondly, OCC is more suitable when the other class is absent, improperly sampled, or not correctly specified. The imbalance issue is common in medical data [28] and in our case, the KID2 dataset has 1,778 normal images while the number of polyp images is only 44.

There are four main categories for an OCC system. First, the density-based methods, such as mixture of Gaussian and Parzen density estimations, get the distribution of a target class [29]. However, these types of methods need a high number of examples. Second, the reconstruction-based methods, such as an autoencoder in neural networks, attempt to capture the structure of a target class [30]. Third, the boundary-based methods, such as SVDD and minimum spanning tree, predict the boundary enclosing the target class [31]. The key challenge of these approaches is to find the optimal size of the model enclosing the given training instances, since choosing too small a one will lead to an overtrained model, while too large a one contributes to an unnecessary inclusion of outliers into the target class. Finally, the ensemble-based methods, such as an ensemble of one-class classification, that cover whole target space are based on clustering [32].

In this paper, we used a hybrid method to use strengths from all OCC groups. The SVDD is used as a boundary-based component, the variational auto-encoder is used for feature extraction as a reconstruction-based component, and an ensemble of SVDDs is generated based on clustered data using Ordering Points to Identify the Clustering Structure (OPTICS) [33] as an ensemble-based component to build a stronger OCC. More description of the combination of OPTICS and SVDD is provided in Supplementary Material 1. Concisely, the extracted features from the autoencoders are clustered using OPTICS and SVDD is applied on each cluster. Outputs of each OCC are ensembled using XGBoost classifier [34] to predict the final class. XGBoost classifier works based on a boosting mechanism, wherein subsequent models are attempting to correct the error of the previous one by giving higher weight to inaccurate predictions. In this case, the weighted average of all models was used as the final model. The pseudo code of the whole process is provided in Figure 4.

```
The pseudo code of the training procedure
Results: A model that can detect anomaly
Step 1: Data preparation:
| Shuffle data
| Stratified partitioning 80% of the data for training, 20% for test set.
Step 2: Feature extraction:
| For each class:
|     Train an autoencoder model based on loss function equation (2)
|     Extract features for images in the class
Step 3: Clustering and one-class classification:
| For features obtained from each autoencoder:
|     Until termination criteria of ACO are met, run the following steps:
|     Generate (next) ant's population for MinPts
|     Clustering features using OPTICS
|     For each cluster:
|         Generate (next) ant's population for σ and C
|         Train SVDD on cluster using gradient decent
|     Optimize F1-score on training data
Step 4: Ensemble of classifiers
| Feed all training features to one-class classifiers
| Use output of each one-class classifiers as input feature for training XGBoost
```

**Figure 4**. The pseudo code for training the proposed method. More descriptions about hyperparameters defined in the pseudo code are available in Supplementary Material 1.

### 2.2.4 Evaluation

Model interpretation refers to ways that humans use to understand the behavior and expectations of a system [35]. For understanding what latent features models are extracting from images, two different approaches are taken. First, the heatmap from the last layer of the encoder is calculated. Because of the dense layers that are used as the estimating average and the standard deviation after the MaxPooling layer, interpretating the heatmap is difficult due to this transformation. Nevertheless, the heatmap may convey the information passed through the network. For calculating the heatmap, the MaxPooling from all 32 filters in the last layer is aggregated to show the important regions detected by the encoder. Then the average and standard deviations of aggregation are fused with the input image to show important regions.

t-Distributed Stochastic Neighbor Embedding (t-SNE) is a dimensionality reduction technique that is ideally suited for the visualization of high-dimensional data [36]. In addition to the heatmap, the extracted latent feature from the model is visualized using t-SNE for better interpretation of the trained models. All training samples are fed into the model and the t-SNE of the latent features are calculated with perplexity 30.

To have consistent results, the whole algorithm ran 5 times with shuffling. Each time 80% of data is used as the training set, 20% is used as the test set. Finally, the average and standard deviation of performance metrics on the test set is reported. The systematic performance metrics used in this research are shown in Table 2 [37]. The reason for choosing the F1-score as the fitness function of optimization is that it is robust to imbalanced data problems and can just be skewed from one way [38], while selecting other objective or fitness function introduced bias towards the majority [39]. All methods and analysis are performed using a computer with Intel Core i9-9900 3.6 GHz CPU and 16 GB of RAM without GPU.

**TABLE 2**. Indices for measuring performance

| Parameter | Definition | |
|---|---|---|
| | 2-class | Multiclass (Macro Average) |
| Recall | $\dfrac{TP}{TP+FN}$ | $\dfrac{1}{l}\sum_{i=1}^{l}\dfrac{TP_i}{TP_i+FN_i}$ |
| Precision | $\dfrac{TP}{TP+FP}$ | $\dfrac{1}{l}\sum_{i=1}^{l}\dfrac{TP_i}{TP_i+FP_i}$ |
| F1-score | $\dfrac{2 \times Recall \times Precision}{Recall + Precision}$ | $\dfrac{1}{l}\sum_{i=1}^{l}\dfrac{2 \times Recall_i \times Precision_i}{Recall_i + Precision_i}$ |
| Accuracy | $\dfrac{TP+TN}{TP+TN+FP+FN}$ | $\dfrac{\sum_{i=1}^{l}TP_i}{TP+TN+FN+FP}$ |

*True positive ($TP_i$): images belong to class i correctly classified; True positive ($TN_i$): images do not belong to class i correctly classified; false positive ($FP_i$): images do not belong to class i classified incorrectly as images in class i; true negative (FN): images belong to class i incorrectly identified as images in class.*

3. **RESULTS**

   **3.1. DEEP FEATURE EXTRACTION**

   For the proposed KID2 dataset, which is a four-class classification task, four autoencoder models are trained. By setting $\lambda_c = \lambda_d = 0$ and $\lambda_k = 0.5$, a simple autoencoder is acquired. For understanding latent features extracted from images, four examples of heatmaps from different classes are obtained from the last layer of the encoder are provided in Figure 5. Besides, the t-SNE plots of deep features is depicted.

   The heatmap should have places on the (8, 8) matrix where maximum values appear (pink color). Places where values change in different channels can also be informative about various extracted features based on standard deviation (green color). The white color shows the positions that have both maximum and standard deviation between channels. All the colored positions show the parts where the model focused its attention.

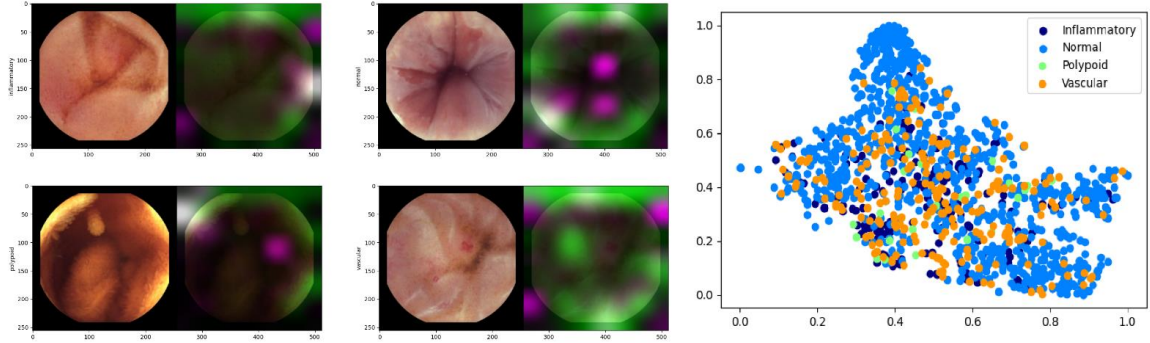

**Figure 5.** The heat map and t-SNE plot acquired from last layer and latent features, respectively, by training a simple autoencoder. The green color represents regions with high standard deviations; the pink regions have the maximum value in filters; the white regions are portions that have both standard deviations and maximum values. Abnormal regions are not captured by the encoder and t-SNE shows the difficulty of distinguishing classes.

The t-SNE plot in Figure 5 illustrates the difficulty of the problem, where a simple autoencoder could not distinguish between features. It is worth noting that t-SNE does not show real distance between samples and two points from different parts of a plot may be very close in term of distance. The t-SNE algorithm adjusts its definition of distance to the regional density variations. Hence, dense clusters naturally grow while sparse clusters contract, balancing cluster sizes.

Next, $N$ autoencoder models are trained with $\lambda_c = \lambda_d = \lambda_k = 0.25$, according to the loss function equation (1). Figure 6 shows the result as a t-SNE and heatmap acquired after 100 epochs. It is clear from the colored region that the encoder could put more attention on the region of the image that has the anomaly. For example, in a polypoid image, the white color shows the polyp, while in a vascular image the white color encounters the red portion, which shows the vascularity. For an inflammatory image, most of the attention is on desired part. For a normal image, the attention is uniformly distributed. Similarly, the t-SNE plots show that each model could separate the desired features from other classes.

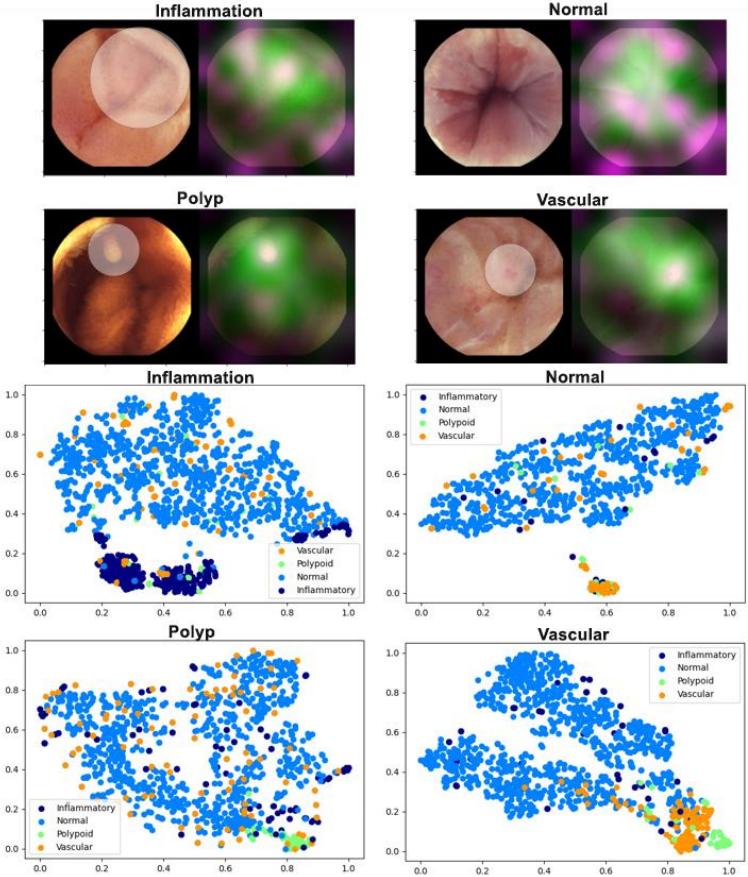

**Figure 6.** The heat map and t-SNE plot acquired from last layer and latent features, respectively, by training four autoencoders on the KID2 dataset. Each heatmap and corresponding t-SNE are obtained from one autoencoder that was purposely trained on specific data. In the heatmap, the green color represents regions with high standard deviations; the pink regions have the maximum value in filters; the white regions are portions that have both standard deviations and maximum values. Abnormal regions, which are captured by the encoder, are visible with more attention in heatmap plots and are highlighted by transparent white circles on original image.

### 3.2. CLUSTERING AND CLASSIFICATION

Figure 7 shows an example of clustered features for the inflammation class. As mentioned in the previous section, the t-SNE plot does not show real distance between samples and two points from different parts of the plot may be very close in terms of distance. However, the hierarchical structure of the clusters can be obtained using a reachability distance (RD) plot. It is a two-dimensional plot with the OPTICS-processed point ordering on the x-axis and the RD on the y-axis. Clusters appear as valleys in the RD plot because their points have a low RD to their nearest neighbor.

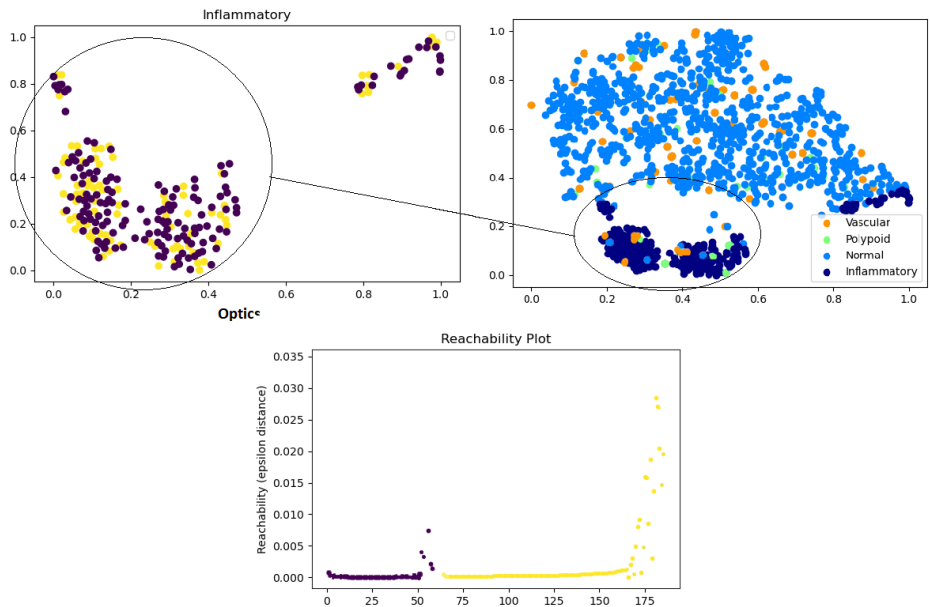

**Figure 7.** The reachability distance plot obtained by OPTICS, with MinPts 7, on the output from an autoencoder trained on the inflammatory class. Note that the t-SNE plot does not show real distance between samples, therefore close points in the t-SNE plot can be from different clusters as OPTICS clustered close points from t-SNE in different classes.

Two different tasks, including binary anomaly detection and multiclass anomaly detection, are investigated. For detecting anomalies in binary format, a group of SVDDs, where each one belongs to a normal class, is trained on clusters acquired by OPTICS on normal data. The training procedure is performed 5 times, and each time, the data is shuffled. The average and standard deviation of accuracy, precision, recall and the F1-score are 94.9 ± 0.3 %, 94.9 ± 0.3 %, 97.7 ± 0.3 % and 96.3 ± 0.2 %, respectively. The boxplot of precision recall and the F1-score for all five runs is depicted in Figure 8.

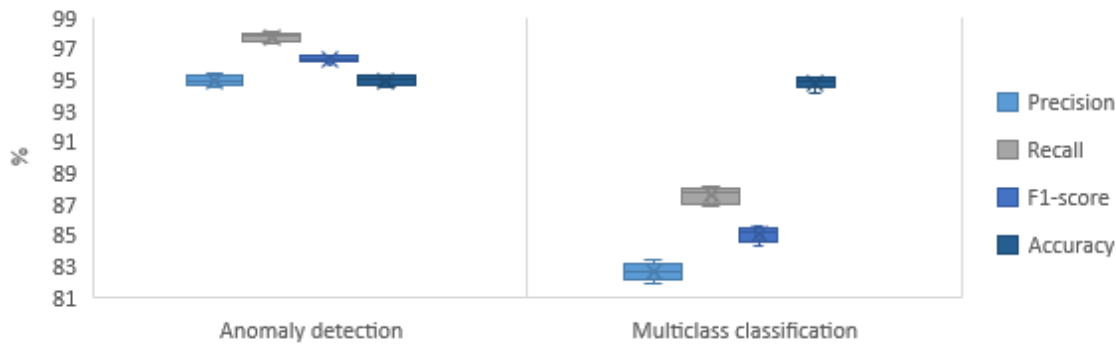

**Figure 8**. The performance of the proposed method for anomaly detection and multiclass classification run 5 times with shuffling training data.

Similarly, for classifying images to specific anomaly categories, the training procedure is performed 5 times and each time the data is shuffled. The average and standard deviation of accuracy, precision, recall and the F1-score are 94.8 ± 0.4 %, 82.6 ± 0.5 %, 87.6 ± 0.5 % and 85.0 ± 0.4 %, respectively. The boxplot of precision recall and the F1-score for all five runs is depicted in Figure 8. Figure 9 shows the performance of the proposed method for each class in the multiclass classification problem.

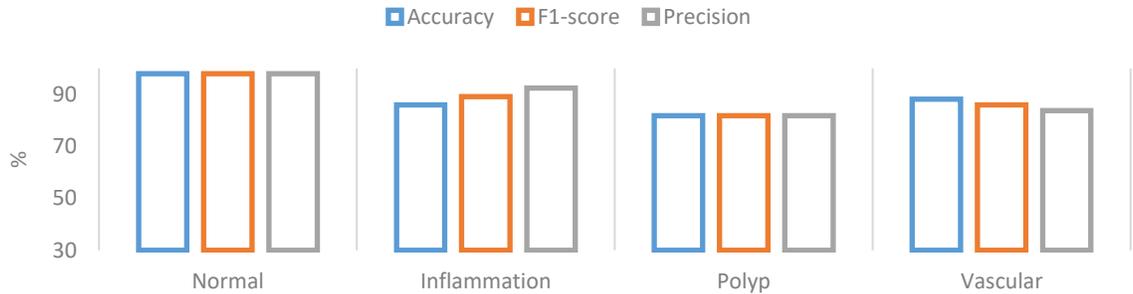

**Figure 9** The average performance of the proposed method for each class on multiclass classification. Recall and accuracy of single class are same.

## 4. DISCUSSION

A multistep approach is presented for multiclass anomaly classification, which can work well on an imbalanced dataset. The method is based on training multiple autoencoders to extract feature from images, wherein each autoencoder is more sensitive to one of the classes. The resulting features are clustered, and a OCC is performed for each cluster to describe the distribution of data portions. Each step of the whole algorithm plays a significant role in performance. To clarify, different ablation studies are presented.

There have been several works on computer-aided decision support schemes to improve diagnostic accuracy for anomaly detection. Table 3 provides a comparison between several relevant techniques from the literature and the proposed method.

**TABLE 3.** COMPARISON BETWEEN THE PROPOSED METHOD AND RECENT APPROACHES.

| Article Cited | Year | Detected output | Binary or Multiclass(#classes) | Dataset | Validation strategy (number of frames for total /test) | Method | Performance % |
|---|---|---|---|---|---|---|---|
| [8] | 2013 | Polyp | Binary | PA [9] | hold-out (270/30) | Inpainting diffusion algorithm combined with energy map | 84.2 Accuracy |
| [40] | 2014 | Polyp | Binary | Private | hold-out (294/65) | Variance of color channels with SVM | 90.7 Accuracy |
| [13] | 2017 | Polyp | Binary | Private | Five times hold-out with shuffling (1930/150) | Transfer learning for feature extraction and SVM for classification | 87.0 F1-score 85.9 Accuracy |
| [10] | 2019 | Polyp | Binary | Private | External dataset (73) | KoloPol APDS software | 85.3 Accuracy |
| [11] | 2020 | Polyp | Binary | Private | hold-out (2684/338) | GI-Genius, Medtronic software | 82.0 accuracy |
| [15] | 2016 | Inflammatory | Binary | KID1 [14] | hold-out (400/54) | CNN | 90.2 Accuracy |
| [41] | 2016 | Inflammatory | Binary | Private | External dataset (231) | Edge and texture analysis | 84.0 Accuracy |
| [17] | 2019 | Anomaly | Binary | KID2 | 10-fold (2352) | Look-Behind Fully CNN | 88.2 Accuracy |
| [16] | 2020 | Anomaly | Binary | KID1 and KID2 | 5-fold (2448) | Random forest-based ensemble classifier with fractal features | 85.0 Accuracy 84.0 F1-score |
| [18] | 2018 | Anomaly | Binary | KID2 | 10-fold (2352) | multi-scale and multi-label CNN | 90.0 AUC |

| | | Anomaly (erythema, blood, ulcer, erosion, and polyp) | | | | | |
|---|---|---|---|---|---|---|---|
| [20] | 2014 | Anomaly (erythema, blood, ulcer, erosion, and polyp) | Multiclass (6) | Private | 10-fold (1750) | Dictionary learning with KNN | 91.1 Accuracy |
| [19] | 2020 | Anomaly (14 different anomalies) | Multiclass (14) | Private | hold-out (28304/14152) | Residual LSTM | 55.0 F1-score |
| **Proposed method** | 2021 | Anomaly (inflammatory, polyp, vascular) | Multiclass (4) | KID2 | Five times hold-out with shuffling (2352/ 470) | Optimized OCC | **85.0 F1-score 94.8 Accuracy** |
| **Proposed method** | 2021 | Anomaly | Binary | KID2 | Five times hold-out with shuffling (2352/ 470) | Optimized OCC | **96.3 F1-score 94.9 Accuracy** |

PA: Publicly available; NA: Not available; hold-out (total size/ test size); External dataset (total size); k-fold (total size).

A fair comparison should consider many factors; therefore, it is hard to say that the proposed method is better than any other technique. Most of the methods did not train a model for multiclass classification. Methods that have multiclass classification have less accuracy or a lower F1-score than the proposed method and are only applied on a private dataset, so we could not apply the proposed algorithm to it. One of the reasons that Mohammed *et al.* [19] have a lower F1-score than the proposed method is because they had a higher number of classes and, as mentioned before, an increasing number of classes makes the problem harder. Although the proposed method could get a higher overall performance score, some other methods are better when trained specifically on one anomaly. For example, according to Figure 9, the polyp detection accuracy is 81.8%, while all other research achieved better results. Nevertheless, the polyp class was the minority class in this research, and the reported results were achieved in a highly imbalanced dataset. For inflammatory detection, the proposed method could achieve 86% accuracy, which is better than Ševo *et al.* [41], but lower than Georgakopoulos *et al.* [15]. Concisely, the proposed method could achieve the best overall F1-score and accuracy for binary and multiclass anomaly detection.

It is worth mentioning that OCC cannot be superior to multiclass and binary approaches when data is balanced, standardized, and well-sampled. It is clear that binary and multiclass approaches have access to counterexamples, which help them estimate the best separation plane. However, OCC approaches can capture the nature of their target class and cover decision space sufficiently; hence they are robust to novelties and have a good generalization ability. Moreover, it is likely that OCC kernel-based methods are able to find a compact description of the data that was enclosed in an atomic hypersphere owing to kernel mapping. Krawczyk *et al.* [42] confirmed all of the above points and showed that OCC could outperform binary classification on seven datasets, mainly because the datasets were highly imbalanced. This demonstrates that using OCC to decompose data is a promising research direction. However, determining why OCC does so well is not always clear.

Further investigation and testing of the proposed method on other datasets with different imbalance issues is required for proving that the proposed method is a robust and general tool. Furthermore, instead of a multi-step framework, the algorithm should be design somehow to be differentiable. This helps to design an end-to-end deep learning algorithm for classification based on gradient descent. Currently, the clustering part, which is based on OPTICS, is not differentiable. Another point is that kernels mostly used for support vector data descriptors are Gaussian, linear, and polynomial, so investigating other kernels such as genetic kernel [43] has the potential to improve the performance.

## 5. CONCLUSION

A novel multiclass classification algorithm for anomaly detection is introduced. The proposed method can work well when the dataset is poorly sampled or imbalanced and is extensible to any number of classes. Features are extracted using multiple autoencoders where each one is trained to cluster the latent features of one class. For this purpose, a new loss function is defined for reconstruction loss and Kullback-Leibler (KL) divergence to increase the inter-class and decrease the intra-class Euclidean distance of features. The proposed algorithm for feature extraction is compared with a conventional autoencoder by plotting the t-SNE and the heatmap acquired from the networks. It showed that the proposed feature extraction scheme is powerful in extracting discriminant features, and it helped the neural network to focus better on regions of interest. Extracted features are clustered into small parts, and for each portion a one-class classification algorithm is trained. The outcome of OCCs is combined and ensembled using XGBoost for predicting anomalies. The proposed method is applied on the KID2 dataset for binary and multiclass anomaly detection and achieved 96.3 ± 0.2% for detecting binarized anomalies and an 85.0 ± 0.4% F1-score for classifying types of anomalies. Although other classifiers can be applied to extracted features, results showed that using OCC gives a better and more robust performance. The proposed method is compared with state-of-the-art and transfer-learning approaches that use cross-entropy loss for image classification. Results showed that the proposed method has the potential to obtain superior performance compared to other multiclass classification models, especially when data is poorly sampled or insufficient in size for calibrating a transfer-learning model.

### DECELERATIONS

#### FUNDING

The Authors would like to acknowledge funding from Natural Sciences and Engineering Research Council of Canada (NSERC) to support the work.

#### AVAILABILITY OF DATA AND MATERIAL

All data used in this research is publicly available at [KID Datasets – mdss.uth.gr](KID Datasets – mdss.uth.gr).

#### CODE AVAILABILITY

[antecessor/MultipleAutoencoderFS: Endoscopy anomaly detection using multiple autoencoder feature selection (github.com)](antecessor/MultipleAutoencoderFS)

#### CONFLICT OF INTEREST

The Authors declares that there is no conflict of interest and competing interest.